\documentclass[11pt, singlecolumn, copyright, logo]{googledeepmind}

\usepackage[authoryear, sort&compress, round]{natbib}
\usepackage{hyperref}
\usepackage{url}
\usepackage{enumitem}
\usepackage{placeins}
\usepackage{float}
\usepackage{graphicx}
\usepackage{bm}
\usepackage{subcaption}
\usepackage{listings}
\usepackage{xcolor}
\usepackage{epstopdf}

\title{Context Structure Reshapes the Representational Geometry of Language Models}

\keywords{representations, in context learning, prediction, language models}

\reportnumber{} % Leave blank if n/a

\renewcommand{\today}{}

\author[1]{Eghbal A. Hosseini}
\author[1]{Yuxuan Li}
\author[1]{Yasaman Bahri}
\author[2]{Declan Campbell}
\author[1]{Andrew Kyle Lampinen}

\affil[1]{Google DeepMind}
\affil[2]{Princeton Neuroscience Institute, Princeton University}

\correspondingauthor{ehos@google.com}

\begin{abstract}
Large Language Models (LLMs) have been shown to organize the representations of input sequences into straighter neural trajectories in their deep layers, which has been hypothesized to facilitate next-token prediction via linear extrapolation. Language models can also adapt to diverse tasks and learn new structure in context, and recent work has shown that this in-context learning (ICL) can be reflected in representational changes. Here we bring these two lines of research together to explore whether representation straightening occurs \emph{within} a context during ICL. %ICL is reflected in representation straightening. 
We measure representational straightening in Gemma 2 models across a diverse set of in-context tasks, and uncover a dichotomy in how LLMs' representations change in context. In continual prediction settings (e.g., natural language, grid world traversal tasks) we observe that increasing context increases the straightness of neural sequence trajectories, which is correlated with improvement in model prediction. Conversely, in structured prediction settings (e.g., few-shot tasks), straightening is inconsistent---it is only present in phases of the task with explicit structure (e.g., repeating a template), but vanishes elsewhere. These results suggest that ICL is not a monolithic process. Instead, we propose that LLMs function like a Swiss Army knife: depending on task structure, the LLM dynamically selects between strategies, only some of which yield representational straightening.

\end{abstract}

\begin{document}

\maketitle

\section{Introduction}

Large language models can leverage the information in their context to adapt to new tasks, role play personas, retrieve specific knowledge, and generalize to novel settings\citep{reynolds2021prompt,Brown2020-or,Shanahan2023-rv,lampinen2023passive}. The structure of the context can shape the model's adaptability toward new behaviors or even negative traits \citep{Chan2022-jy}. A central challenge in mechanistic interpretability is thus understanding the relationship between the structure of context and how LLMs reshape their internal representations in response to adapt their behavior \citep{akyurek2022learning,raventos2023pretraining,wang2023label}.

Existing research has broadly explored this problem along two complementary axes. The first focuses on the \textit{data distribution}, examining how input structure influences model behavior \citep{Chan2022-jy,Lampinen2022-sz,Chen2024-ba,Agarwal2024-ag,Lee2025-rn,Shanahan2023-rv,lampinen2023passive}. The second focuses on \textit{mechanistic explanations}, identifying specific circuits and representational motifs underlying emergent behaviors \citep{Olsson2022-cv,Singh2024-kr,Park2024-he,Li2025-eq,Von-Oswald2023-ti,Park2024-uk}. Here, we examine how these lines of work intersect: exploring how different types of data structure in context change the models' representational geometry. 

We address this question by interrogating the evolution of representational structure during ICL. We build upon the framework of \citet{Hosseini2023-rj} in language modeling, which modeled the sequence encoding process as a neural trajectory within the representation space (see also \cite{Henaff2019-zc} for visual perception). Multiple lines of work have investigated how artificial or biological systems can compress the past information for predicting the future \citep{le1991mpeg,radford2018improving,oord2018representation,tishby2000information,palmer2015predictive,pmlr-v119-henaff20a,eysenbach2024inference,ziarko2025contrastive,Henaff2019-zc}. They hypothesize that LLMs transform the previous input to make future input more predictable, by structuring the representation of inputs to be more straight and thus predictable via \emph{linear extrapolation}. Straighter trajectories facilitate next-token prediction. A similar phenomenon was previously observed in the brain's visual system, and human visual perception \citep{Henaff2019-zc,Henaff2021-sk} and finds support in recent information-theoretic \citep{Skean2025-gz} and mechanistic perspectives \citep{Lubana2025-cm}. We hypothesize that language models might similarly reorganize their representations \emph{during} in-context learning and therefore form straighter trajectories in representation space.

Given this geometric perspective, we investigated representational geometry during ICL behavior in Gemma-2 models \citep{Gemma-Team2024-fe}. Drawing on the extensive literature on data structures in in-context learning, we evaluated a spectrum of data structures, including natural language \citep{Paperno2016-fs}, grid worlds with controlled latent structure \citep{Park2024-uk}, and few-shot learning benchmarks \citep{Todd2023-xc,Li2025-eq,Srivastava2022-jw}. 

\begin{figure}[t!]
\centering
\includegraphics{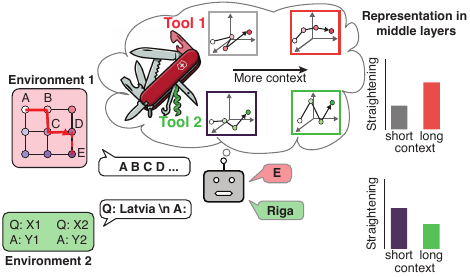}
\caption{Models select computational strategies based on task structure, which influence the representational structure and geometrical measures over them } \label{fig:message}
\end{figure}

Our analysis reveals that representational geometry depends on the task context. %a nuanced view the "straightening hypothesis."
While we observe consistent representational straightening in natural language and structured grid tasks, suggesting a flattening of representational manifolds, we find a \emph{dissociation} in few-shot learning and question-answering tasks. In these settings, the changes in straightening do not correlate with the model's ability to solve the task. This dissociation suggests that LLMs do not rely on a universal mechanism for all forms of context. Instead, models may learn a \emph{toolkit} of distinct mechanisms---akin to a Swiss Army knife---wherein context evokes the computational mechanism most appropriate for the task at hand (Fig.\ref{fig:message}).

% add a section on contribution 

\section{Methods} \label{sec:Methods}

\subsection{Tasks} \label{sec:datasets}

To investigate the diversity of representational mechanisms in In-Context Learning (ICL), we consider three distinct classes of tasks. These tasks were chosen to probe the model's capacity for ICL across different cognitive demands: (1) capturing long-range dependencies in natural language, (2) inducing latent structure from synthetic sequences via graph traversal, and (3) performing algorithmic and semantic reasoning via few-shot prompts and riddles.

\paragraph{Natural language long-range dependencies task:} We used the LAMBADA dataset \citep{Paperno2016-fs}, a collection of narrative passages where the target word (the final word of the passage) is predictable only when given the broader context, and not from the immediate local context (the last sentence). This dataset emphasizes long-range contextual dependencies, providing a naturalistic setting to examine how context shapes representational geometry to support next-token predictions.

\paragraph{Grid world tasks:} Following \citet{Park2024-uk}, we created synthetic grid world tasks to test the model's ability to infer latent graph structures from linear sequences of tokens. These tasks map a graph of nodes with transitions to a sequence of tokens. We consider two levels of complexity:

\begin{itemize}[topsep=0pt, itemsep=2pt]
    \item \textit{One level of abstraction (direct mapping):} Similar to \citep{Park2024-uk}, we construct a graph with 36 nodes arranged on a $6\times6$ non-periodic lattice. Each node is randomly assigned a unique, common English word (guaranteed to be a single token for the model; examples: apple, bird, car, egg, house, milk, plane, opera, box). Contextual sequences consist of random walks of variable length generated over this graph, with total sequence lengths up to 1024 tokens.
    
    \item \textit{Two levels of abstraction (hierarchical mapping):} We construct a graph with 16 latent nodes on a $4\times4$ non-periodic lattice. Each latent node is associated with 4 "child" observations (words), selected to be semantically similar to one another (categories: nature, animals, food, household objects, transportation, abstract concepts, body parts, places, roles, accessories, calendar, materials, hardware, electronics, geographical features, emotions). While in-context sequences are generated via a random walk over \textit{latent} parent nodes, the model observes only the corresponding child words (sampled uniformly from the active parent node). This forces the model to infer the underlying latent trajectory from noisy observations. We generated sequences with lengths up to 2048 tokens.
\end{itemize}

\textit{Evaluation Protocol:} To evaluate in-context learning, we constructed a test set of short, 5-token walks. These sequences represent valid continuations of the context but are novel; neither the exact test sequences nor their constituent 4-token sub-sequences appeared in the context window. 

We probed performance under three distinct conditions to assess the impact of context length and structure:
\begin{enumerate}[topsep=0pt, itemsep=0pt]
    \item \textit{Short context:} Test sequences were inserted between token positions 5 and 64 of the random walk.
    \item \textit{Long context:} Test sequences were inserted at random positions within the final 64 tokens of the context window.
    \item \textit{0-shot context (Hierarchical only):} Transitions along specific edges between children of neighboring latent nodes were excluded from the context but presented at test time. Success in this condition requires the model to infer the existence of the latent edge connecting the parent nodes despite never observing that specific child-to-child transition.
\end{enumerate}

For each task and condition, we generated a dataset of 200 unique sequences (context + test sequence).

\paragraph{Few-shot learning task:} We selected a subset of tasks from \citet{Todd2023-xc,Li2025-eq}, spanning both semantic processing and algorithmic reasoning. All tasks are formatted as standard few-shot prompts consisting of alternating "Q: [Input]" and "A: [Output]" pairs. This setting explicitly tests the model's ability to map inputs to outputs based on discrete examples rather than continuous narrative flow. For each task we select 100 8-shot examples ( 8 in-context examples + 1 test example) from a larger set of examples, and extracted model representations. 

\paragraph{Riddle task:} To evaluate ICL in a task requiring semantic disambiguation, we used the Riddle benchmark from BIG-bench \citep{Srivastava2022-jw}. Unlike the standard few-shot tasks, riddles retain a natural language structure while having the Q\&A format. Each query is followed by a 5-item multiple-choice set and an "A:" prompt. We analyze changes in representational geometry by contrasting 0-shot performance with 8-shot performance. The original dataset contained 98 unique examples, which we used for the 0-shot case. To create the 8 shot dataset, we prepend each riddle with a random combination of 8 other riddles, and repeated it twice per riddle, resulting in 196 8-shot riddle examples. 

\begin{figure}[t!]
\centering
\includegraphics{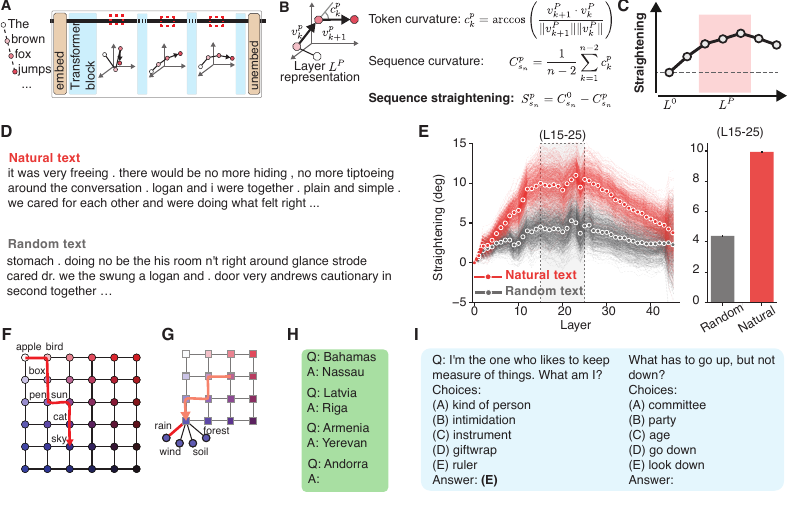}
\caption{Straightening in natural language context, and experimental design(A) Each transformer layer modifies the sequence representation, leading to changes in geometry. (B) Within each layer (ex. $L^p$) we first computed token curvature between all triplets of adjacent tokens in a sequence and took their average to get a sequence curvature. To measure how representation of a sequence changes as it gets transformed by each layer, we computed sequence straightening as the difference between the sequence curvature in first and subsequent transformer layers. 
(C) Changes in straightening are tracked across layers, focusing on the middle layers. (D) We compare natural versus random text to measure straightening induced by context. (E) Models show substantial straightening in middle layers (L15--25) for natural text compared to random. (F--H) We investigate geometry in diverse ICL settings: (F,G) Latent structure induction in Grid Worlds, (H) Few-shot learning, and (I) Semantic reasoning in Riddles} \label{fig:setup}
\end{figure}

\subsection{Models}
Our analysis primarily utilizes the open-weight Gemma-2-27B model \citep{Gemma-Team2024-fe}. We selected the pre-trained (base) version to isolate representational structures emerging from self-supervised text training, minimizing the confounding factors of reinforcement learning from human feedback (RLHF) and other types of post-training, given that recent work has established the representational changes at these subsequent stages \citep{li2025tracingrepresentationgeometrylanguage}.

\subsection{Behavioral Measures}

\paragraph{Grid world task Evaluation.} To quantify the model's understanding of the latent graph structure, we evaluate the logits assigned to valid transitions. For a given token in the test sequence, we compare the logits of the ground-truth "neighbor" nodes (valid transitions in the graph) against "non-neighbor" nodes. We define success as the model assigning higher aggregate logits to the set of neighbors compared to non-neighbor nodes.

\paragraph{Few-Shot Learning task Evaluation.} In few-shot tasks, we evaluate the model's performance by sampling the response token(s) immediately following the final "A:" prompt. Accuracy is determined by exact match with the ground truth answer.

\subsection{Geometric Measures}

We employ four complementary metrics to characterize the geometry of the model's internal representations across layers and context length.

\paragraph{Straightening.} Following \citet{Hosseini2023-rj,Henaff2019-zc}, we estimate the local curvature of the neural trajectory by examining the angle between consecutive transition vectors. Let $x^p_1, x^p_2, \dots, x^p_n$ denote the activations in the residual stream at layer $L_p$ for a sequence of $n$ tokens. We define the transition vector at step $k$ as $v^p_k = x^p_{k+1} - x^p_k$. The local curvature $c^p_k$ is defined as the angle between adjacent transitions:
$$ c^p_k = \arccos\left(\frac{v^p_{k+1} \cdot v^p_{k}}{\|v^p_{k+1}\| \, \|v^p_{k}\|}\right) $$
A value of $0$ indicates a straight line (collinear transitions), while higher values indicate a more curved path. We report the sequence-level curvature as the average over the token window:
$$ C^p_{s_n} = \frac{1}{n-2} \sum_{k=1}^{n-2} c^p_k $$
We define straightening of each sequence as the difference between curvature at the first transformer layer $L_0$ and $L_p$.

$$S^p_{s_n} = C^0_{s_n} - C^p_{s_n}$$

\paragraph{Menger Straightening.} To capture a complementary geometric measure of the trajectory, we compute the Menger curvature \citep{Hahlomaa2008-ct}. Using the transition vectors $v^p_k$ defined above, we first normalize them to unit length: $\hat{v}^p_k = v^p_k / \|v^p_k\|$. For any three consecutive points on this normalized path, we define a triangle with side lengths $a, b, c$. Since adjacent sides are unit vectors, $a=b=1$, and $c$ is the Euclidean distance between the endpoints of the triplet. The local Menger curvature $\kappa^p_k$ is the reciprocal of the radius of the circumcircle passing through these three points:
$$ \kappa^p_k = \frac{4K}{abc} = \frac{4\sqrt{s(s-a)(s-b)(s-c)}}{abc} $$
where $K$ is the area of the triangle and $s$ is the semi-perimeter. As before we compute sequence Menger curvature as the average of token level curvatures:
$$ \mathcal{K}^p_{s_n} = \frac{1}{n-2} \sum_{k=1}^{n-2} \kappa^p_k $$
We report the Menger straightening as $\mathcal{K}^0_{s_n}$- $\mathcal{K}^p_{s_n}$ across the sequence.

\paragraph{Effective Dimensionality.} To quantify the volume of the representation space occupied by the trajectory, we used the Participation Ratio. Let $X^p \in \mathbb{R}^{n \times d}$ be the matrix of centered activations for the sequence at layer $L_p$. We perform Principal Component Analysis (PCA) on $X^p$ to obtain the eigenvalues $\lambda_1, \dots, \lambda_d$ of the covariance matrix. The effective dimensionality is given by:
$$ ED^p = \frac{(\sum_{i=1}^d \lambda_i)^2}{\sum_{i=1}^d \lambda_i^2} $$
This metric provides a continuous estimate of the number of principal components required to capture the variance of the neural trajectory.

\paragraph{Elongation.} We further characterize the anisotropy of the trajectory using the Elongation metric. Using the eigenvalues from the PCA described above, where $\lambda_1$ and $\lambda_2$ represent the variances along the first and second principal components respectively, we define elongation $E^p$ as:
$$ E^p = 1 - \frac{\lambda_2}{\lambda_1} $$
This metric ranges from 0 to 1, where a value close to 1 indicates that the trajectory is maximally elongated (lying predominantly along a single linear manifold), while a value close to 0 implies a more isotropic distribution across the principal plane.

\section{Results}

\subsection{Representational straightening in natural language task}

We begin by establishing a geometric baseline for the Gemma-2-27B model in a natural language setting (Fig.~\ref{fig:setup}D). This task, a subset of the LAMBADA dataset, requires integration of long-range dependencies for successful prediction. We measure \emph{representational straightening} of token sequences ($n_{sequences}=600$) across all transformer layers (Fig \ref{fig:setup}A,B). As a control, we also measured straightening in a "random text" condition where the tokens were shuffled within each sequence, destroying semantic structure while preserving the token distribution.

The geometric profile of natural text shows a progressive increase in straightening that begins in early layers and peaks in the middle layers (Layers 15–25, Fig.~\ref{fig:setup}E). This suggests that the model actively linearizes the neural sequence trajectory, untangling the input manifold of natural language into a flatter geometry. In the final layers, straightness gradually decreases. We suspect that this reduction reflects the unembedding phase, where the high-dimensional representation must collapse back into the geometry of the vocabulary space to generate discrete next token predictions.

Critically, the straightening diminishes in the random text control condition (Fig.~\ref{fig:setup}E). This confirms that representational straightening is not an artifact of the architecture, but reflects the construction of predictable token sequences. These results replicate and extend prior findings, which found similar phenomena across diverse autoregressive architectures and throughout training \citep{Hosseini2023-rj,Skean2025-gz}, and further establishes straightening as a proxy for predictive processing in natural language.

Having established this baseline, we examine whether the \emph{straightening} is present across different forms of in-context learning. Specifically, we analyze two categories of tasks (Fig.~\ref{fig:setup}F--I):

\begin{itemize}[topsep=0pt, itemsep=2pt]
    \item \textbf{Continual prediction tasks} (Grid Worlds, Fig.~\ref{fig:setup}F,G): tasks where the model must infer hidden graph structure from observed sequences.
    \item \textbf{Structured prediction tasks} (Fig.~\ref{fig:setup}H,I): tasks requiring the application of specific input-output mappings (e.g., Country $\rightarrow$ Capital) based on discrete examples, as well as tasks requiring ambiguity resolution and semantic manipulation to retrieve answers.
\end{itemize}

\begin{figure}[t!]
\centering
\includegraphics{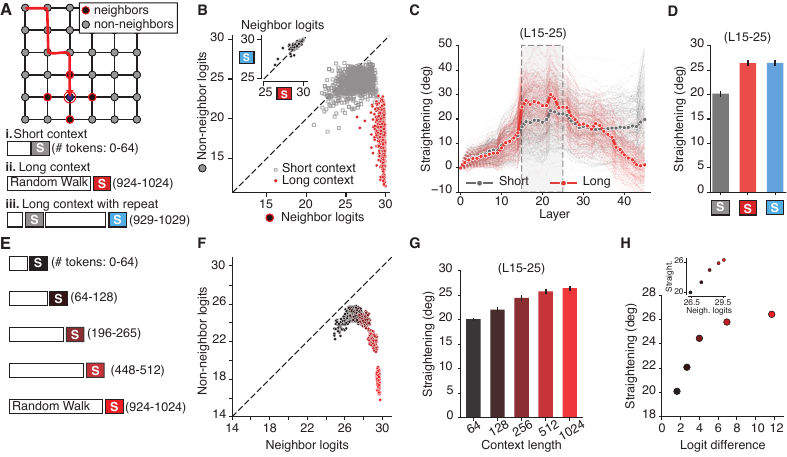}

\caption{Contextual adaptation in continual prediction task is reflected in straightening. (A) We probed the model's knowledge of the grid using three test conditions: (i) Short context, (ii) Long context, and (iii) Long context with repeat. (B) Logit values for neighbor versus non-neighbor tokens. While these values are similar in the short context, neighbor logits are significantly higher in the long context. (Inset) Neighbor logits are similar between the standard long context and the 'repeat' condition. (C) Straightening of target sequences across layers; long context exhibits substantially greater straightening in the middle layers. (D) Average straightening (L15-25) for short, long, and repeat conditions. (E) Experimental setup with progressively longer context windows to interrogate learning dynamics. Same color scheme is used in panels F-H. (F) Divergence of neighbor and non-neighbor logits as context length increases. (G) Corresponding increase in straightening in middle layers with longer context. (H) Relationship between the behavioral logit difference (neighbor minus non-neighbor) and straightening. (Inset) relationship between raw neighbor logits and straightening. } \label{fig:basegrid}
\end{figure}

\subsection{Geometric structure in grid world task}

We next investigate the model's ability to learn latent structure in the "Grid World" tasks. We provide the model with random walks on a $6\times6$ lattice (36 nodes), followed immediately by a short, unobserved 5-token test walk. To differentiate contextual learning from memorization, we created three walk conditions (Fig.~\ref{fig:basegrid}A): (i) \textit{short context} (early in the sequence), (ii) \textit{long context} (late in the sequence), and (iii) \textit{long context with repeat}, where the test sequence appeared early in context exactly once (between token positions 5--64). This third condition specifically tests whether the model prediction relies on a "copying" mechanism or a generalized understanding of the graph structure \citep{Olsson2022-cv}.

The model successfully learns the underlying graph structure with longer contexts. We quantified this by comparing the logits assigned to valid graph neighbors versus invalid non-neighbors at each step of the test sequence (Fig.~\ref{fig:basegrid}A, top). In the short context condition, the assigned logit values for neighbors and non-neighbors are similar and cluster near the unity line ($y=x$), indicating the model's lack of knowledge about the underlying graph (Fig.~\ref{fig:basegrid}B). However, in the long context condition, we observe a distinct pattern: the logit values for neighbors increase while non-neighbors decrease, resulting in a downward shift from the unity line. An independent samples t-test over logit differences (x-y) confirmed that the distributional shift is robust ($t=-122.28, p<0.001,n_{samples}=1000$). Critically, the presence of the test sequence earlier in the context (Condition iii) yields a similar logit profile to the standard Long Context (Fig.~\ref{fig:basegrid}B, inset). While an independent samples t-test between neighbor logits showed a difference ($t=-2.23, p=0.026, n_{samples}=1000$) this effect size was very small (Cohen's $d=-0.10$, i.e, two sets differ by $0.1$ standard deviation). This suggests that the model is not relying solely on a copy mechanism, but is instead querying a learned representation of the graph structure.

This behavioral shift is mirrored by changes in representational straightening. Comparing the short and long context conditions, we find that the accumulation of context leads to an increase in \emph{straightening} in the middle layers (Layers 15–25) (Fig.~\ref{fig:basegrid}C, D). The neural trajectory over test sequences becomes significantly more straight in the long context condition (independent samples t-test: $t=-7.46,p<0.001, d=0.75$). Consistent with the behavioral results, the "repeat" condition shows comparable straightening to the standard long context (independent samples t-test: $t=0.07,p=0.95,d=0.007$) (Fig.~\ref{fig:basegrid}D).

To understand the dynamics of task learning, we analyzed the changes across five context lengths (64 to 1024 tokens, Fig.~\ref{fig:basegrid}E). We observe a tightly coupled change between logit patterns and straightening. As context length increases, the separation between neighbor and non-neighbor logits grows (Fig.~\ref{fig:basegrid}F), mirrored by a consistent increase in straightening (Fig.~\ref{fig:basegrid}G). We quantified this relationship by correlating the average straightening at each context length with the logit difference (neighbor minus non-neighbor logits). While this relationship is monotonic, it is non-linear (Pearson's $r=0.87$, Fig.~\ref{fig:basegrid}H). However, the relationship between straightening and the \textit{neighbor} logits is much more linear ( Pearson's $r=0.99$, Fig.~\ref{fig:basegrid}H, inset), suggesting that straightening is directly tied to the changes in logits for neighbors.

\begin{figure}[t!]
\centering
\includegraphics{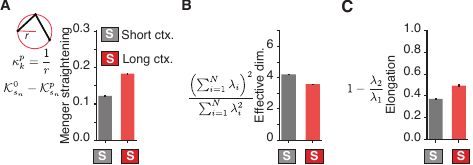}

\caption{Straightening generalizes to other representational measures.(A) Changes in target sequence representation measured via Menger straightening; long context induces higher straightening. (B) Effective dimensionality of target sequences; long context sequences exhibit lower effective dimensionality. (C) Elongation of target sequences; long context trajectories are more anisotropic, biased toward the first eigenvector direction. } \label{fig:add_measure}
\end{figure}

Straightening is based on a linearity assumption and collapses all residual dimensions into one measure. We therefore tested tested three complementary geometric measures to see whether they similarly point to linearity in representations(Fig.~\ref{fig:add_measure}).Here we focused on the same set of tokens sequences as straightening measure( the 7-token :2 prefix tokens+ 5 test sequence) measured changes from short to long context in the middle layers (L15-25):

\paragraph{Menger Straightening.} We computed the normalized Menger straightening (where $0$=random, $2$=collinear) \citep{Hahlomaa2008-ct}. Consistent with straightening metric, the \textit{long context} induces higher Menger straightness in middle layers (independent samples t-test: $t=-13.91,p<0.001, d=1.39$) (Fig.~\ref{fig:add_measure}A).

\paragraph{Effective Dimensionality.} We estimated the effective dimensionality of the 7-token trajectory (test sequence + 2 prefix tokens). As context increases, the effective dimensionality decreases (Fig.~\ref{fig:add_measure}B), indicating that the representation contracts onto a lower-dimensional manifold (independent samples t-test: $t=13.75, p<0.001, d=1.37$).

\paragraph{Elongation.} We measured the anisotropy of the sequence trajectory. \textit{Long context} results in higher elongation (Fig.~\ref{fig:add_measure}C), indicating that the local trajectory evolves robustly along a single dominant direction (independent samples t-test: $t=-9.27, p<0.001, d=0.93$).

Together, these results suggest that as the model processes longer context, it creates smoother representations that could facilitate prediction of correct transitions.

\subsection{Geometric structure in latent grid world task}

\begin{figure}[t!]
\centering
\includegraphics{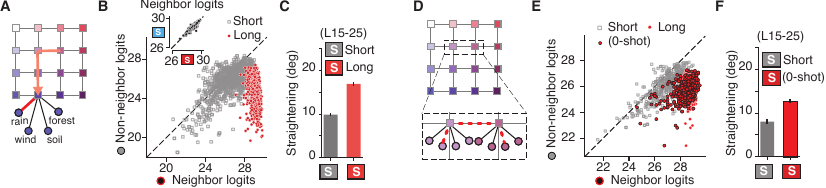}

\caption{Straightening generalizes to learning over latent graphs (A) Latent graph structure: random walks occur on latent nodes, but the model observes only emitted tokens (e.g., "rain"). (B) Neighbor versus non-neighbor logits; long context increases neighbor logits compared to non-neighbors (C) Straightening in middle layers for short and long context. (D) Zero-shot probe: specific transitions between children of connected latent nodes were withheld during context (dashed red line) and introduced only at test time. (E) Neighbor vs. non-neighbor logits for Short, Long, and Long (0-shot) conditions. The 0-shot logits separate effectively, mirroring the fully observed Long context. (F) Straightening for Short vs. Long (0-shot) transitions; straightening is larger for inferred transitions later in context.} \label{fig:latentgrid}
\end{figure}

We next investigate whether the model learns simple surface statistics or infers latent structure in grid world task. We created \textit{latent grids} where random walks occur on a hidden $4\times4$ lattice, but the model observes only "child" tokens emitted by latent nodes, with each latent node emitting 1 of 4 tokens at random(Fig.~\ref{fig:latentgrid}A).

We exposed the model to random walks with similar short/long context format, albeit twice as long (2048 tokens instead of 1024 for long context; short context was still 64 tokens). The model successfully recovers the latent graph structure, and effectively differentiates logits for tokens belonging to latent neighbors versus non-neighbors (Fig.~\ref{fig:latentgrid}B, independent sample t-test comparing short to long context $t=-61.1, p<0.001, d=2.7$; independent sample t-test comparing 1 with 2 repeat $t=0.01, p=0.98, d=0.0$ ). The change in logits is accompanied by the same geometric signature observed previously: an increase in \emph{representational straightening} in the middle layers (Fig.~\ref{fig:latentgrid}C), indicating the formation of a linearized geometry (independent sample t-test $t=-11.62, p<0.001, d=1.16$).

To rule out memorization of token-token pairs, we tested a \emph{0-shot transitions} condition. Within a context, we excluded transitions between specific child tokens of connected latent nodes(Fig.~\ref{fig:latentgrid}D). In test sequences however we inserted these unseen transitions, and tested whether the model can infer these transitions via the learned latent graph. Behaviorally, the model successfully predicts these 0-shot transitions (Fig.~\ref{fig:latentgrid}E, independent sample t-test comparing short vs 0-shot: $t=-31.1,p<0.001,d=2.2$, independent sample t-test comparing 0-shot with long context $t=11.1,p<0.001, d=0.78$ ). Critically, the geometric signature persists: we observe an increase representational straightening for these inferred transitions in long context (Fig.~\ref{fig:latentgrid}F, independent sample t-test: $t=-6.0, p<0.001, d=0.6$). These results confirm that the model learns the underlying latent graph in context, and this learning is geometrically expressed as an increase in straightening of sequence neural trajectories.

\subsection{Geometric structure in few-shot learning tasks}

The tasks examined so far (natural language and grid worlds) have a continual prediction format: the model predicts the next token by integrating context over many previous observation. We next focus on emergent in-context capability of the model in structured prediction tasks and study \emph{few-shot learning}, where the model must induce an algorithmic rule or manipulate semantic knowledge from discrete examples to solve a task \citep{Brown2020-or,Todd2023-xc, Li2025-eq}. We used a set of tasks involving factual manipulation (e.g., "Country-Capital") and function learning with distinct semantic/algorithmic processing (e.g., "Position of fruit among animals", "Choose last of five") \citep{Li2025-eq, Todd2023-xc}. These tasks probe the model's emergent ability to perform logical/semantic operations in context.

\begin{figure}[t!]
\centering
\includegraphics{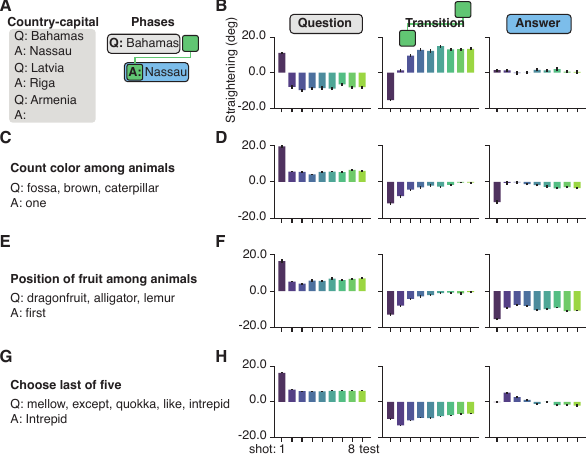}

\caption{Representation of few-shot learning task is not reflected in straightening. (A) Phasic decomposition of few-shot learning task examples into Question, Transition, and Answer. (B) "Country-Capital" task: The question phase shows decrease in straightening after the first shot, while the transition phase shows an increase in straightening with more examples. The answer phase shows no significant change. (C-F) Similar dissociation dynamics are observed in "Count color among animals", and "Position of fruit among animals" tasks. (G-H) "Choose last of five" task: The question phase shows reduction in straightening, while transition and answer phases exhibit bi-phasic dynamics, initially reversing straightening direction before recovering with additional examples. } \label{fig:taskvectors}
\end{figure}

To track the changes in the geometry of representation, we decomposed each few-shot example into three phases (Fig.~\ref{fig:taskvectors}A):
\begin{enumerate}[topsep=0pt, itemsep=2pt]
    \item \textbf{Question phase ($Q$):} The input stimuli (e.g., "Q: Latvia").
    \item \textbf{Transition phase ($T$):} The structural formatting tokens (e.g., "\textbackslash n A:").
    \item \textbf{Answer phase ($A$):} The target output (e.g., "Riga").
\end{enumerate}
This decomposition allows us to track changes in representation that are driven by fixed structural templates from those driven by task learning. We tracked the evolution of representational straightening across these phases from the first shot ($k=1$) to the final test shot, ($test$), focusing on sequences where the model successfully generated the correct answer.

Unlike the uniform straightening observed in grid worlds, few-shot learning reveals a \emph{dissociation} between the structural and learning components of the task (Fig.~\ref{fig:taskvectors}B, see suppl.Fig.~\ref{fig:sup_fewshot_layers} for straightening results across all layers). The question phase shows a sharp \textit{drop} in straightening after the first example. In the country-capital task for example, we find a significant difference in straightening between the 0-shot and 1-shot conditions (independent samples t-test: $t=14.96, p<0.001, d=2.18$). However, this effect disappears for subsequent examples, suggesting that once the model recognizes the task structure, it no longer processes the query tokens as a linear continuation. Conversely, the transition phase exhibits a robust, continuous increase in straightening as the number of examples increases ( one-way ANOVA on effect of shot: $F=152.7,p<0.001$). This trend mirrors our findings in natural language and grid world tasks, suggesting that the model treats the rigid formatting template ("\textbackslash n A:") as predictable continuation and linearizes it. Critically, the answer phase, where the actual task solution is generated, does \textit{not} exhibit the progressive straightening observed in previous tasks and degree of straightening of the answer trajectory remains unchanged (one way ANOVA on the effect of shot: $F=0.67, p=0.72$).

We observed a similar dissociation across a broader set of algorithmic and semantic tasks (Fig.~\ref{fig:taskvectors}C-H, see suppl.Fig.~\ref{fig:sup_fewshot_layers} for additional examples). In tasks requiring semantic and algorithmic manipulation (e.g., "Count color among animals") or pure algorithmic manipulation (e.g., "Choose last of five"), we observe a similar trend: the transition phase straightens consistently, while the question phase shows sharp change in the first example and answer phase does not show a consistent pattern of change.

This finding challenges the hypothesis that representational straightening is a universal geometrical feature across diverse ICL tasks structures. Instead, it suggests that straightening emerges in tasks with continual prediction format, while the \emph{computational core} of the task with structural mapping (mapping Question $\to$ Answer) relies on a distinct mechanism that is not reflected in straightening.

\subsection{Geometric structure in riddle task}

While synthetic few-shot tasks reveal a geometric dissociation, their structure is artificial. To test naturalistic reasoning, we examine the \emph{Riddle Benchmark} \citep{Srivastava2022-jw} . We contrasted performance in a 0-shot baseline versus an 8-shot context to isolate the effect of in-context learning.

\begin{figure}[t!]
\centering
\includegraphics{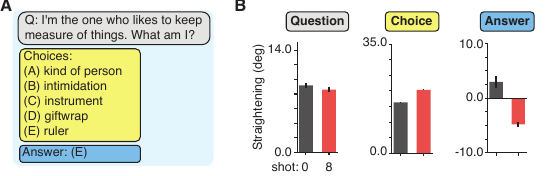}

\caption{Representation of few-shot riddle task is not reflected in straightening. (A) We decomposed each example of the few-shot riddle task into three phases: Question, Choice, and Answer. (B) Changes in straightening from the 1st to the 8th example in context. While the question phase shows a reduction in straightening, the choice phase shows an increase in straightening. The answer phase, on the other hand, shows a substantial reduction in straightening from the 1st to the 8th example in context.} \label{fig:riddle}
\end{figure}

To track the changes in representation geometry, we decomposed each riddle interaction into three distinct phases, in order to separate narrative flow from structural formatting (Fig.~\ref{fig:riddle}A):
\begin{enumerate}[topsep=0pt, itemsep=2pt]
    \item \textbf{Question phase ($Q$):} natural language clues (e.g., "I'm the one who...").
    \item \textbf{Choice phase ($C$):} a list of options (e.g., "(A) person...").
    \item \textbf{Answer phase ($A$):} the target prediction (e.g., "Answer: (E)").
\end{enumerate}

The pattern of straightening across question, choice and answer phases of the riddle task confirm a geometric \emph{dissociation} observed in few-shot learning (Fig.~\ref{fig:riddle}B, see suppl.Fig.~\ref{fig:sup_riddle} for straightening results across all layers). The question phase shows a slight non-significant \textit{decrease} in straightening with context (independent samples t-test: $t=1.24, p=0.21, d=0.18$). Conversely, the choice phase exhibits an increase in straightening, indicating the model is linearizing the choice formatting (independent samples t-test: $t=-12.46, p<0.001, d=1.8$). Critically, the answer phase shows a significant \textit{decrease} in straightening with context (independent samples t-test: $t=6.38,p<0.001, d=1.0$).

These results align with the findings in few-shot learning tasks, confirming that straightening is shaped by the \emph{continual prediction component} of the task, while the \emph{computational core} (resolving semantic ambiguity) relies on a distinct mechanism that is not reflected by straightening.

\FloatBarrier
\section{Discussion}

In this work, we probed the representational geometry of In-Context Learning (ICL) across diverse tasks, identifying a dichotomy in representational strategies. Our findings suggest that LLMs do not rely on a monolithic representational mechanism, but rather a "toolkit" of strategies adapted to the specific structure of the task.

While our results support \textbf{representational straightening} as a signature of learning continual prediction, they also highlight its limitations. In natural language and grid worlds tasks, context accumulation drives the neural trajectory onto a more linear/low-dimensional manifold, in support of \citep{Hosseini2023-rj,Henaff2019-zc}. However, this measure fails to capture representational changes in structured ICL tasks, such as few-shot learning \citep{Todd2023-xc,Li2025-eq}. This suggests the computational toolkit of the LLM is akin to a \emph{Swiss Army knife}: depending on the task, the model reshapes its representation via distinct, task-appropriate mechanisms \citep{Hendel2023-fm, Todd2023-xc}.

These results offer a cautionary perspective for the work on interpretability. While it is useful to build intuition and a representational toolkit for specific task settings, one must test across a diverse set of tasks. It is likely that a single \emph{generalized notion} of representational structure underlying ICL does not exist, but rather that the model relies on a library of computational strategies to maintain flexibility, and that is reflected in a set of representational and geometric signatures (c.f. \citealt{Von-Oswald2023-ti, park2023linear,Park2024-uk}).

Our work can be improved in several ways. We focused on a specific set of geometrical measures, leaving other geometrical/topological features unexplored. Additionally, while we showed a correlation between straightening and behavior in grid world tasks, we did not perform causal interventions. Developing causal interventions that manipulate these geometric features without pushing the model representation off of its natural data manifold remains a key challenge. \citep{Hase2023-ua,McGrath2023-ud}. Finally, we restricted our analysis to the Gemma-2 architecture; expanding this evaluation to a broader range of model families and scales could further refine our observations.

\subsection*{Acknowledgments}

We thank Stephanie Chan, Jay McClelland, Mike Mozer, Katherine Hermann, and Vaishnavh Nagarajan for helpful comments and discussions. We thank Murray Shanahan, and Michael Terry for support.

%\subsection*{Code Availability}
%
%We are working on getting the code for our main benchmarks approved for an open-source release in time for publication. 
%
%\subsection*{Data Availability}
%
%The data used to reproduce the analyses in the main text is available upon request. 

\bibstyle{unsrtnat}
\bibliography{main}

\newpage
\appendix

\setcounter{figure}{0} % Resets the figure counter to 0
\renewcommand{\thefigure}{S\arabic{figure}} % Changes numbering to S1, S2...
\section{Supplemental data}

\begin{figure}[tbh]
\centering
\includegraphics{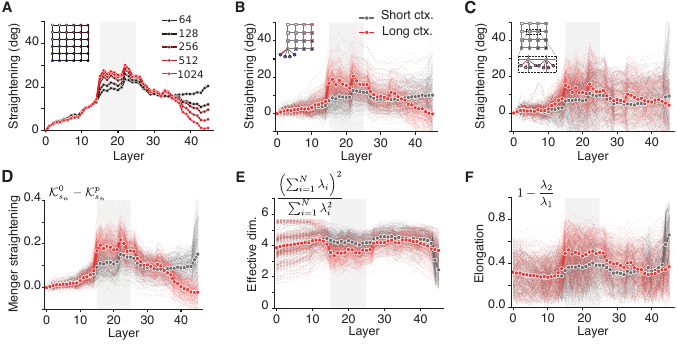}

\caption{(A) Straightening across layers in the grid world task with different context lengths. (B) Straightening across layers in the latent grid world task with short context (maximum 64 tokens) and long context (maximum 2048 tokens). (C) Straightening across layers for zero-shot generalization in the latent grid world task. (D) Menger straightening across layers in the grid world task. (E) Effective dimensionality across layers in the grid world task. (F) Elongation across layers in the grid world task.} \label{fig:sup_grid_layers}
\end{figure}

\begin{figure}[tbh]
\centering
\includegraphics{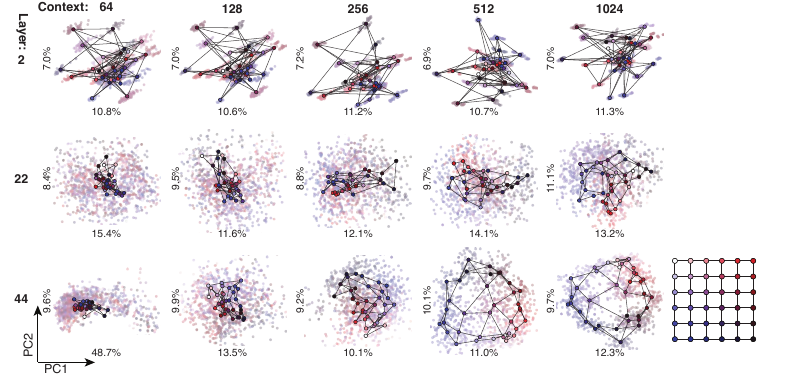}

\caption{Representation of the grid structure at different layers and across contexts in the grid-world task. Each plot shows the PCA of the representation over individual test sequences, with darker points showing the average per node. Edges are overlaid on these average values to reveal the representation of the grid (or lack thereof). The percentage values are the variance explained by PC1 and PC2, respectively. Rows show early (2), middle (22), and late (44) layers in the Gemma 2-27b-it model (with 46 layers), while columns show progressively longer contexts prior to test sequences, from 64 tokens to 1024. While early-layer representation does not change with increasing context, middle and late layers reveal a restructuring of the representation. Among them, the late layer has the closest alignment with the grid structure shown in the bottom-right panel. These results corroborate findings from \citep{Park2024-uk}. } \label{fig:sup_pca}
\end{figure}

\begin{figure}[tbh]
\centering
\includegraphics{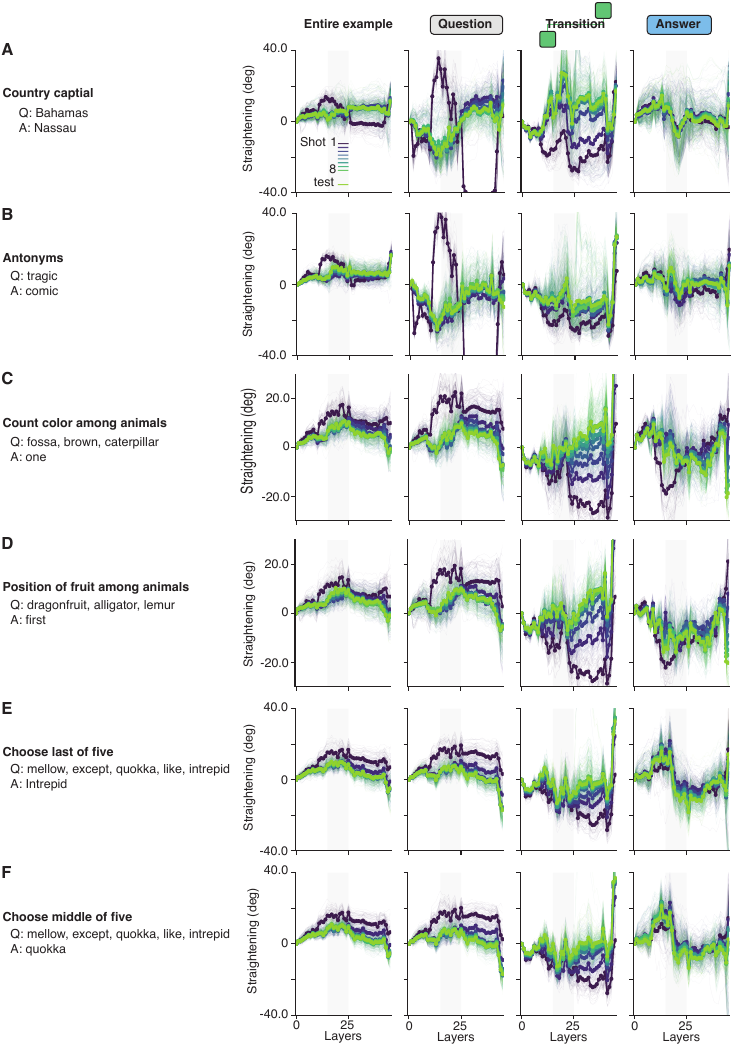}

\caption{(A) Straightening across layers for the country-capital condition in the few-shot learning task. Straightening across all tokens in an example is shown in the first column, while individual phases (Question, Transition, and Answer) are shown in subsequent columns. Colors transition from the 1st example to the 8th, and the test example. The results are shown for cases where the model response is correct on the test. (B) Antonyms task. (C) Counting colors among animals task. (D) Position of fruit among animals task. (E) Choose the last of first words task. (F) Choose the middle of five words task.} \label{fig:sup_fewshot_layers}
\end{figure}

\begin{figure}[tbh]
\centering
\includegraphics{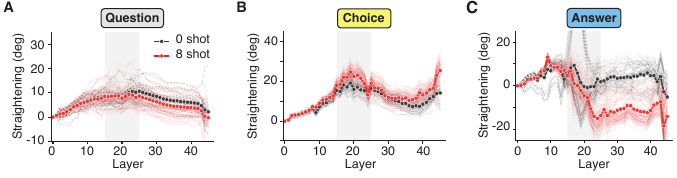}

\caption{(A) Straightening across layers during the question phase of the riddle task, in the 0-shot vs. 8-shot conditions. (B) Same as (A), but for the choice phase. (C) Same as (A), but for the answer phase. } \label{fig:sup_riddle}
\end{figure}

\end{document}